\pdfoutput=1

\documentclass[11pt]{article}
\usepackage{authblk}

\usepackage[final]{acl}

\usepackage{times}
\usepackage{latexsym}

\usepackage[T1]{fontenc}

\usepackage[utf8]{inputenc}

\usepackage{microtype}

\usepackage{inconsolata}

\usepackage{multirow}
\usepackage{multicol}
\usepackage{amsmath}
\usepackage{graphicx}
\usepackage{amssymb}
\usepackage{siunitx}
\usepackage{caption}
\usepackage{subcaption}
\title{Fusion of Domain-Adapted Vision and Language Models for Medical Visual Question Answering}

\author[1]{\bf Cuong Nhat Ha}
\author[2]{\bf Shima Asaadi}
\author[2]{\bf Sanjeev Kumar Karn}
\author[2]{\authorcr \bf Oladimeji Farri}
\author[2]{\bf Tobias Heimann}
\author[1,3]{\bf Thomas Runkler}
\affil[1]{Technische Universität München}
\affil[2]{Digital Technology and Innovation, Siemens Healthineers AG}
\affil[3]{Corporate Technology, Siemens AG}
\affil[ ]{\tt cuong.ha@tum.de}
\affil[ ]{\tt \{shima.asaadi,sanjeev.kumar\_karn,oladimeji.farri,tobias.heimann\}@siemens-healthineers.com}
\affil[ ]{\tt thomas.runkler@siemens.com}

\begin{document}
\maketitle
\begin{abstract}
Vision-language models, while effective in general domains and showing strong performance in diverse multi-modal applications like visual question-answering (VQA), struggle to maintain the same level of effectiveness in more specialized domains, e.g., medical. We propose a medical vision-language model that integrates large vision and language models adapted for the medical domain. This model goes through three stages of parameter-efficient training using three separate biomedical and radiology multi-modal visual and text datasets. The proposed model achieves state-of-the-art performance on the SLAKE 1.0 medical VQA (MedVQA) dataset with an overall accuracy of $87.5$\% and demonstrates strong performance on another MedVQA dataset, VQA-RAD, achieving an overall accuracy of $73.2$\%. 


\end{abstract}

\section{Introduction}
Vision-Language Models (VLM), composed of two key elements - vision models and language models, mainly establish a connection between text-based and image-based modalities. In order to accomplish this fusion, VLMs undergo training using large volumes of text and images. This training process enables them to understand the correlations between visual and textual data, thus equipping them to handle tasks such as Visual Question Answering (VQA).

Vision-language models, such as CLIP \cite{clip} and BLIP-2 \cite{blip2}, have shown impressive performance across various multi-modal applications. Nevertheless, these VLMs have not displayed similar levels of performance when applied to the Medical VQA (MedVQA) task \citep{biomedclip}. 
The complexity of medical questions in MedVQA often requires a deep understanding of medical terminology and image context that may not be adequately captured by a generic VLM. 
Therefore, recent approaches, such as PubMedCLIP \citep{pubmedclip23}, Med-Flamingo \citep{medflamingo}, LLAVA-Med \citep{llavamed}, and Biomed-CLIP \citep{biomedclip} adapt general-domain VLMs to the medical domain by leveraging large datasets containing both medical images and accompanying text, such as ROCO \citep{roco}. 




Moreover, prior approaches, including PubMedCLIP \citep{pubmedclip23} and the models studied by \citet{medvqa_survey}, treated MedVQA as a classification problem, where the models had to choose the correct answer from a predefined set. This approach not only restricts the ability of VLMs to generate free-form responses but also leads to inaccurate evaluation.


In this paper, we first define the MedVQA task as free-text generation, which is considered a more challenging task compared to classification.
Next, we present a novel vision-language model that fuses a domain-specific Large Language Model (LLM) customized for radiology with a vision model designed for biomedical tasks. 
In the proposed vision-language model, all parameters of both the vision and language models remain fixed. 
We propose a parameter-efficient training approach by integrating Low-Rank Adaptation (LoRA) technique \cite{lora} for training the model. 
The frozen domain-adapted models and LoRA training ensure not only stability and consistency during training but also optimize the overall efficiency of the training process.


Our proposed training approach for the trainable parameters consists of three stages: medical concept alignment through the image-captioning task using PMC-OA dataset \cite{pmcclip}, adaptation to the general medical VQA task using the PMC-VQA dataset \citep{pmc-vqa}, and fine-tuning on the radiology task specific training dataset, such as VQA-RAD \citep{vqarad} and SLAKE 1.0-English \citep{slake}.


We conducted evaluations on two public radiology MedVQA evaluation benchmarks, VQA-RAD \citep{vqarad} and SLAKE 1.0 \citep{slake}, to assess the performance improvement achieved by our proposed VLM. Our model outperformed existing models from published works on the SLAKE 1.0 benchmark, achieving an impressive overall accuracy of 87.5\%. Furthermore, our model demonstrated strong performance on the VQA-RAD benchmark, highlighting its effectiveness compared to other published models. Additionally, we conducted a performance comparison between our model and a version that incorporates a general-domain LLM while keeping all other components constant. We observed a big performance improvement with the domain-adapted language model, and thereby demonstrating the advantage of integrating these models into VLMs as a promising approach to address the limitations of adapting general VLMs to domain-intensive applications.



Lastly, in our ablation investigation, we evaluated the effect of our proposed multi-stage training approach and found that it led to a significant 25\% improvement in accuracy compared to directly fine-tuning a general-domain VLM on the downstream MedVQA task. Our analysis underscores the advantages of incorporating a domain-specialized LLM into the VLM architecture and highlights the effectiveness of our proposed training strategy in addressing MedVQA tasks.

Our contributions can be summarized as follows:
\begin{itemize}
    \item We introduce a multi-modal model for MedVQA by fusing a radiology domain-specific decoder-only LLM with a bio-medical vision model within a VLM framework. 
    \item We propose a parameter-efficient three-stage training approach for efficient and effective fusion of a vision encoder and LM.
    \item Our proposed model outperforms the state-of-the-art on the SLAKE 1.0 MedVQA dataset. Furthermore, we thoroughly analyze our model and approach using both quantitative and qualitative methods.
\end{itemize}

The remaining paper is structured as follows.
In Section 2, we provide a detailed description of
the model with its training schema. In Section 3, we
describe and discuss the dataset and experiments. In Section
4, we discuss the related works. Section 5 concludes the study.

\begin{figure*}[h]
\small \centering
\includegraphics[width=\linewidth]{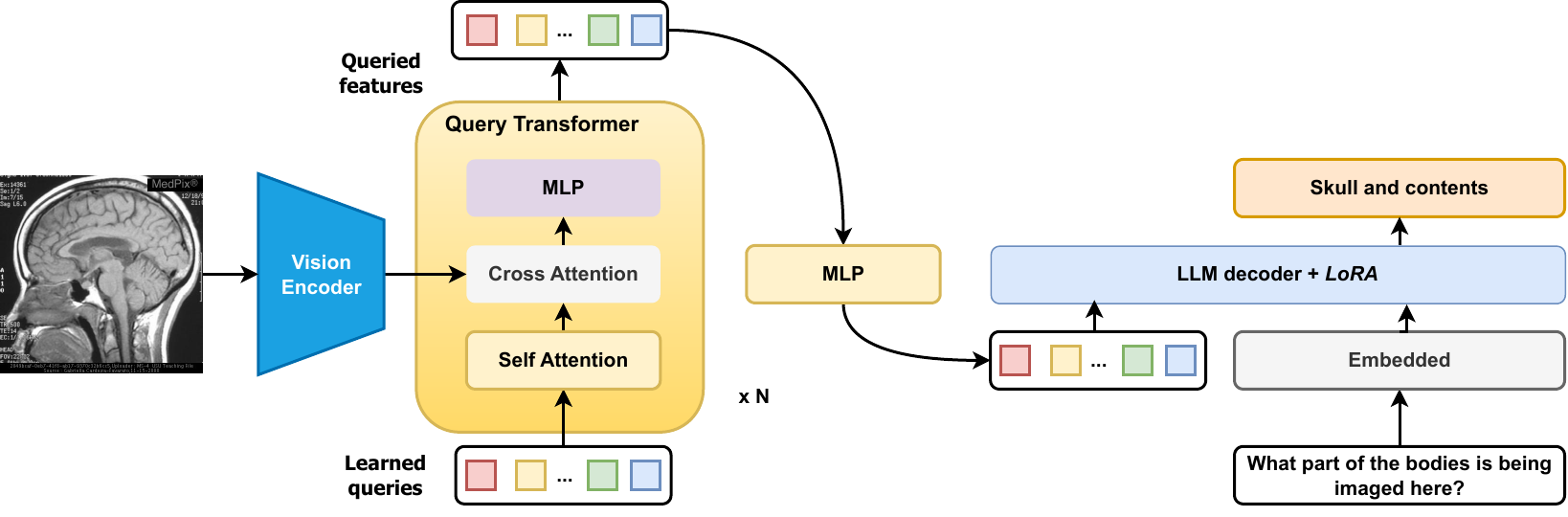}
\caption{Overview of the proposed vision-language (VLM) architecture for MedVQA task. The output from the biomedical-adapted vision encoder component is combined with the input question, processed through a Radiology-adapted Language Model (LLM). Learned queries are initiated from scratch and trained during our proposed alignment training of multi-modal domain adapted models, which includes image-caption pretraining, synthetic biomedical MQA, and MedVQA datasets, all fine-tuned using a parameter efficient LoRA technique.
}
\label{model_arch}
\end{figure*}

\section{Model}
\textbf{Problem Formulation}:
Given a medical image $v_i$ and a natural language question $q_i$, a trained VLM model $\mathcal{M}$ with parameters $\Theta$ generates the answer $a_i$ for the given question as:
\begin{center}
    \label{eq:objective}
    \begin{equation}
            {a}_i = \mathcal{M}(v_i, q_i; \Theta),
    \end{equation}
\end{center}
where ${a}_i$ is the generated answer.
Unlike previous approaches that treat MedVQA as a classification task, where the answer $a_i$ is selected from a predefined set of possible answers $\{\ldots{a}_i,\ldots\}$, our objective is to generate an open-ended answer $a_i$ instead. 

Figure \ref{model_arch} shows our VLM model architecture. Our model includes a vision encoder that takes in the image $v_i \in \mathbb{R}^{H \times W \times C}$, where $H$, $W$, and $C$ denote the height, width, and channels of the image, respectively. It outputs the encoded image $e(v) \in \mathbb{R}^{n \times m}$, with an embedding size of $m$ and $n$ number of patches.

In our VLM model, the fusion module serves the purpose of mapping the encoded vision features $e(v)$ to the embedding space of the LLM. This module acts as a bridge between the vision encoder and the LLM. Taking inspiration from BLIP-2 \citep{blip2}, we employ a learnable query transformer architecture as the fusion module. Its primary function is to extract a predetermined set of features from the output of the vision encoder. The parameters of this module are randomly initialized.

The query transformer output is transformed using a multi-layer perceptron network to match the embedding size of the LLM, resulting in $e(v)\;' \in \mathbb{R}^{d}$. These projected features are then combined with the embedded input text $e(q) \in \mathbb{R}^d$ and fed into the LLM to generate the desired output.




In order to explore the potential benefits of incorporating radiology domain-adapted Language and vision models in MedVQA tasks that involve radiology images, questions, and answers, we utilize decoder-only transformer models as the LLM module. More specifically, we leverage RadBloomz-7b \citep{radbloomz}, which is a radiology domain adaptation of Bloomz-7b1 \citep{bloomz}. 

The RadBloomz-7b model has been continuously pre-trained using the MIMIC-IV radiology reports dataset \citep{mimic4} and has demonstrated exceptional performance on the radiology report summarization task, surpassing other models on the MIMIC-III \citep{mimic3}, MIMIC-CXR \citep{mimic-cxr}, and CheXpert \citep{chexpert} summarization datasets. We argue that RadBloomz-7b offers a highly powerful foundation model and brings valuable advantages to downstream MedVQA tasks.

To investigate the potential advantages of integrating domain-specific vision models into MedVQA, we utilize the vision encoder models from PMC-CLIP \cite{pmcclip} and BiomedCLIP \cite{biomedclip}. These models have demonstrated notable performance enhancements in multi-modal medical tasks, including question-answering. By employing these models, we not only have access to two different pre-trained vision models but also have the opportunity to explore two distinct architectures: ResNet50 \cite{resnet} from PMC-CLIP \cite{pmcclip} and Vision Transformer (ViT) from BiomedCLIP \cite{biomedclip}.

In our model, the vision encoder and LLM remain as pre-trained models with frozen parameters. Instead, we propose using the Low-Rank Adaptation (LoRA) technique \citep{lora} on the pre-trained LLM to align it with the downstream MedVQA task. 


\subsection{Training Approach}
\label{training_approach}

Our training approach comprises three main stages, with the first two stages considered as pre-training and the final stage as fine-tuning. The loss function employed in all training stages is the sum of negative log-likelihoods of the correct next token in a given text sequence across all time stages as: 
\vspace{-1em}
\begin{equation}
    L(\Theta) = - \sum_{t=1}^{T} \log p(a_t | v, q, a_{1:t-1}; \Theta),
\label{eq_loss}
\end{equation} where $\Theta$ is the trainable model parameters, $T$ is the length of the ground-truth answer, and $p(\cdot)$ represents the probability of generating the $t$-th token in the answer sequence given the input image $v$, the question $q$, and the previous tokens in the answer sequence $a_{1:t-1}$.

\textbf{Pre-Training Stage 1: Medical concept alignment}: 
This stage is framed as a medical image caption prediction task, where the model predicts the next token in the caption given an input image. 
The loss function is accordingly defined as:
\begin{equation}
\label{eq:loss_caption}
    L(\Theta) = - \sum_{t=1}^{T} \log p(c_t | v, c_{1:t-1}; \Theta),
\end{equation} where $c_{t-1}$ and $c_t$ are the caption tokens at time $t-1$ and $t$, respectively, and $v$ is the input image. 

This stage serves two purposes: bridging the gap between the vision encoder model and language model, and pre-training the randomly initialized fusion module to align medical concepts with visual content. This integration enables the fusion module to understand medical concepts in images and align visual information with textual descriptions. We utilize a training strategy called Image-grounded Text Generation (ITG) in this stage, which is inspired by BLIP-2 \citep{blip2}. However, unlike BLIP-2, we train the introduced LoRA parameters of the LLM.


\textbf{Pre-Training Stage 2: General medical visual question answering}
To build an effective MedVQA model, we rely on the PMC-VQA dataset \citet{pmc-vqa}. This dataset encompasses a diverse collection of medical images across multiple modalities, including X-ray, CT, MRI, and microscopy. It also features a wide range of questions that cover various aspects of medical images. By training the model using this dataset, we expose it to a rich variety of medical scenarios, fostering the development of broad knowledge and generalization in the medical field. The loss function is the same as in Equation \ref{eq_loss}.


We utilized the second version of the PMC-VQA dataset for our training process, which is approximately $186,033$ image-associated questions and answers.


\textbf{Training Stage 3: Downstream task finetuning}
In the final stage, we fine-tune the model by utilizing the training split of two publicly available MedVQA benchmarks: VQA-RAD \cite{vqarad} and SLAKE 1.0-English \citep{slake}. This process helps us further refine the model's performance. The loss function during this stage remains the same as in Equation \ref{eq_loss}. 

\section{Experiments}

\subsection{Experiment setup}
Our objective is to evaluate how well the proposed method performs in answering questions related to medical visual content. To do this, we conduct experiments and compare its performance with the following baseline VLMs.

\begin{itemize}
\item \textbf{BiomedCLIP} \cite{biomedclip}. This biomedical domain adapted vision-language foundation model is pretrained on PMC-15M, which is a dataset consisting of 15 million image-caption pairs extracted from PubMed Central. The model is trained using contrastive learning techniques. Additionally, we consider this model as one of the domain-adapted vision model for our fusion experiments. We make use of the vision component ViT-Base-patch16-224 variant, which has a patch size of $16\times16$. We refer to this variant as \textbf{"BiomedCLIP ViT"}.

\item \textbf{PMC-CLIP}. Inspired by CLIP \citep{clip}, \citet{pmcclip} combine image-text contrastive loss with masked language modeling loss from BERT to train a new model called PMC-CLIP. To pre-train their VLM, \citet{pmcclip} employ the PMC-OA dataset, consisting of $1.6$M image-caption pairs. They combine ResNet50 \citep{resnet} as the vision module and PubmedBERT \citep{pubmedbert} as the language module. Additionally, a 4-layer transformer is trained as the fusion module. Like BiomedCLIP, we utilize the ResNet50 model from PMC-CLIP as a domain-adapted vision model. This variant is referred to as \textbf{"PMC-CLIP ResNet"}.


\item \textbf{MUMC}. \citet{li2023mumc} propose a novel vision language pre-training approach. They use masked image and text encoding with uni-modal and multi-modal contrastive losses on image and text encoders, along with image and text features. They also introduce a masked image strategy for data augmentation by randomly masking image patches during pre-training. For downstream tasks, they incorporate transformer-based decoder layers to generate answers and fine-tune the model using the masked language modeling objective on VQA datasets.

\item \textbf{PubMedCLIP} \citet{pubmedclip23} present PubmedCLIP, a fine-tuned version of CLIP for the medical domain. It is trained on image-text pairs from PubMed articles. The authors explore the impact of incorporating PubmedCLIP as a pre-trained vision encoder in two MedVQA methods. They further fine-tune these models using public MedVQA benchmarks. Due to the inclusion of text encoders, the training and evaluation of MedVQA are structured as a multi-label classification task rather than a free-form generation task.

\item \textbf{MedVInT-TD} \citet{pmc-vqa} propose a generative-based VLM that integrates visual information from vision encoders, such as ResNet from PMC-CLIP \citep{pmcclip}, with large language models, such as PMC-LLaMA-7B \citep{pmcllama} as decoder-only models. They pretrain their model using PMC-OA on the image-captioning task. Then, they introduce a large-scale medical multi-modal question-answering dataset, PMC-VQA, with which their proposed model is instruction tuned. We selected this model for comparison as it's directly comparable to ours, given its similar use of a decoder-only LLM.
\end{itemize}

\subsection{Datasets}
The pre-training process for aligning medical concepts involves two stages. In the first stage, the PMC-OA dataset \cite{pmcclip}, containing 1.64 million image-caption pairs, is used. In the second stage, the version 2 of the PMC-VQA dataset \cite{pmc-vqa}, encompassing approximately 186,033 visual question-answer pairs, is utilized. In the third stage, we utilize the training split of VQA-RAD \cite{vqarad} and SLAKE 1.0-English \cite{slake} datasets for the downstream fine-tuning tasks, as they are the most popular public benchmarks in the radiology domain. For additional information, please refer to Table \ref{tab:datasets} in the Appendix section \ref{data_apn}. In both fine-tuning datasets, questions are categorized as either closed-ended or open-ended. Closed-ended questions are multiple-choice questions with a limited set of answers, such as "yes/no" questions. Open-ended questions contain free-form answers.

\subsection{Training and Evaluation}
\begin{table*}
\small \centering
\begin{tabular}{@{}l@{}ll|ccc|ccc}
\\ \hline
& & & & SLAKE 1.0 & & & VQA-RAD &    \\
Model & VE & LM & Overall  & Closed  & Open & Overall  & Closed  & Open \\
\hline
Ours & BiomedCLIP ViT        &  RadBloomz-7b  & \textbf{87.5}  & \textbf{92.1} & \textbf{84.5}  & 73.2 & 83.5 & 57.5\\ \hline
Ours & PMC-CLIP ResNet50   & RadBloomz-7b & 82.5 & 88.5 & 78.6 & 67.6 & 79.4 & 49.7 \\ \hline
\begin{tabular}{@{}l@{}}MedVInT-TD \\ \citep{pmc-vqa} \end{tabular} & &  & 85.2 & 86.3 & 84.5 & \textbf{81.6} & \textbf{86.8} & \textbf{73.7} \\ \hline
\begin{tabular}{@{}l@{}}Biomed-CLIP  \\ \citep{biomedclip}    \end{tabular} &  & & 86.1 & 88.9 & 84.3 & 72.7 & 76.5 & 67.0 \\ \hline
\begin{tabular}{@{}l@{}}PubMedCLIP \\ \citep{pubmedclip23}   \end{tabular} & & & 80.1 &  82.5 & 78.4 &  72.1 & 80.0 & 60.1  \\  \hline
\begin{tabular}{@{}l@{}}MUMC \\ \citep{li2023mumc}   \end{tabular}     & & & 84.9  & - & - &  79.2  & 84.2 & 71.5  \\  \hline
\begin{tabular}{@{}l@{}}PMC-CLIP \\ \citep{pmcclip}   \end{tabular}    & & & 84.3  & 88.0 & 81.9 &  77.6 & 84.0 & 67.0  \\  \hline
\end{tabular}
\caption{Accuracy (\%) results of VLMs on SLAKE 1.0-English and VQA-RAD datasets. Performance on open-ended and closed-ended questions as well as overall performance are reported. VE represents vision encoder.}
\label{vqa_rad_slake_acc}
\end{table*}

We train our model for 3 epochs in the first stage of aligning medical concepts with an initial learning rate of $3e-4$. For the second stage of pre-training, we trained the model for 10 epochs with a learning rate of $1e-5$. Finally, we fine-tuned the model on MedVQA benchmarks for 100 epochs, using a learning rate of $2e-5$.

For all training stages, we employed the AdamW optimizer \citep{adamw} with a cosine annealing schedule. The training batch size was set to $256$ for pre-training and $16$ for fine-tuning. All training processes were conducted on 4 A100-40GB GPUs. To optimize our training procedures, we integrated the DeepSpeed \citep{deepspeed} acceleration strategy along with Automatic Mixed Precision (AMP) \citep{mix_precision} techniques.

To evaluate the performance on VQA-RAD and SLAKE 1.0-English, we measure the accuracy metric. We further analyze the results by distinguishing between open-ended and closed-ended questions, allowing for a detailed assessment of the model's performance across different question types. 

In our approach to the MedVQA task, we adopt the method proposed by \citet{pmcllama}, which treats it as free-form text generation. We identify the answer in the list of all possible answers from the training split of each dataset that is most similar to the answer generated by our model. We then compare this selected answer to the ground truth. To achieve this comparison, we make use of Python's difflib library.\footnote{\href{https://docs.python.org/3/library/difflib.html}{https://docs.python.org/3/library/difflib.html}}

\subsection{Results and Analysis}
\begin{table*}
\small\centering
\begin{tabular}{@{}ll|ccc|ccc}
\hline
 & & & SLAKE 1.0 & & & VQA-RAD &    \\
VE & LM & Overall  & Closed  & Open & Overall  & Closed  & Open \\
\hline
\multirow{2}{*}{BiomedCLIP ViT } & Bloomz-7b1 & 80.0 & 86.8 & 75.7 & 68.3 & 80.9 & 49.2 \\
& Radbloomz-7b & \textbf{87.5}  & \textbf{92.1} & \textbf{84.5}  & \textbf{73.2} & \textbf{83.5} & \textbf{57.5} \\ \hline
\multirow{2}{*}{PMC-CLIP ResNet }        & Bloomz-7b1 &  80.5 & 87.5 &	76.0 & 65.2 & 77.9 & 45.8  \\ 
& Radbloomz-7b & \textbf{82.5} &  \textbf{88.5} & \textbf{78.6} & \textbf{67.6}  & \textbf{79.4} & \textbf{49.7}  \\ \hline
\end{tabular}
\caption{The table compares the accuracy (\%) between a VLM with a radiology-adapted RadBloomz-7b LM and a general-domain Bloomz-7b1 LM, using the SLAKE 1.0-English and VQA-RAD datasets. Results for open-ended, closed-ended, and overall performance are included, with experiments conducted separately using two pretrained vision encoders (VE).}
\label{vision_comp}
\end{table*}

The results of our proposed model can be seen in Table \ref{vqa_rad_slake_acc}. Its evident that our BiomedCLIP-RadBloomz-7b model achieves state-of-the-art performance on SLAKE 1.0, with an overall accuracy of $87.5$, surpassing the previous approaches. This model excels particularly in closed-ended questions with accuracy of $92.1$. The results illustrate the advantages of our training strategy and the utilization of a radiology domain-adapted language model in the MedVQA task.

Additionally, when comparing similar experiments where the domain-adapted BioMedCLIP-ViT vision encoder is replaced with PMC-CLIP ResNet, it becomes evident that utilizing BiomedCLIP-ViT results in superior performance on both benchmark datasets. The findings indicate that certain domain-adapted vision encoders, such as BiomedCLIP, possess exceptional capabilities in effectively managing domain-specific knowledge within specific language models like RadBloomz-7b. Also, this successful combination underscores the potential for further research in exploring the fusion of these models.


In the VQA-RAD dataset, our BiomedCLIP-RadBloomz-7b model outperforms PubMedCLIP \citep{pubmedclip23} and Biomed-CLIP \citep{biomedclip} models on the overall accuracy. It also demonstrates competitive performance with existing approaches on closed-ended questions. However, it does not perform as well on open-ended questions, where it falls behind compared to the MedVInt-TD model. We argue that the lower performance on open-ended questions can be attributed to several factors. One key factor is our formulation of the problem as free-form answer generation for both question types, as opposed to the baseline Biomed-CLIP and PubMedCLIP models. This means that our model is not constrained by a predefined set of answers in the training data.


To evaluate the influence of domain adaptation in the VLM, we performed experiments using two  LMs, Bloomz-7b1 and RadBloomz-7b. The comparison results in Table \ref{vision_comp} demonstrate that BiomedCLIP-RadBloomz-7b outperforms its general domain language model counterpart, Bloomz-7b1, on both datasets. There is a noticeable enhancement in overall accuracy on Slake 1.0, with an improvement of 7.5\%. Similarly, on VQA-RAD, there is a significant increase in overall accuracy, with an improvement of 4.9\%. This highlights the significant benefit of employing a domain-adapted language model, specifically RadBloomz-7b, as the backend language model for domain-intensive tasks in VLMs. The model's effectiveness is particularly evident in its performance on open-ended questions, demonstrating an average improvement of 8.5\% in accuracy.


To evaluate the impact of including training of existing parameters in the fusion model, we conducted experiments on VLMs that employed trainable vision encoders. In this regard, we trained the vision encoder parameters alongside other trainable parameters throughout all training stages. Table \ref{train_freeze_comp} shows the results obtained from the VLMs using trainable BiomedCLIP-ViT. The two LMs, Bloomz and RadBloomz, were utilized in the experiments. Notably, the VLM utilizing the specialized-domain RadBloomz-7b achieves better performance with a reduced number of parameters compared to the VLM with a larger set of trainable parameters. We argue that through an optimal fusion of the domain-adapted vision encoder and LM, there is no longer a need to train the vision encoder in our VLM. This results in a lightweight adaptation of the VLM.

\begin{table*}
\centering
\begin{tabular}{ll|ccc}
\hline
VE                          & LM & Overall  & Closed-ended  & Open-ended    \\ \hline
Trained BiomedCLIP ViT         & Bloomz-7b1     &  69.4            & 80.1     & 53.1       \\
Frozen BiomedCLIP ViT        
& Bloomz-7b1     &   68.3  &     80.9 & 49.2  \\ \hline
Trained BiomedCLIP ViT & RadBloomz-7b   & 71.4    & 81.3 & 56.4 \\ 
Frozen BiomedCLIP ViT  & RadBloomz-7b   & 73.2   & 83.5 & 57.5 \\ \hline

\hline
\end{tabular}
\caption{The table provides a comparison of accuracy (\%) between two scenarios on the VQA-RAD dataset: one scenario where the vision encoder of VLMs is trained alongside alignment training, and another where the vision encoder is frozen during training. The table displays performance for open-ended and closed-ended questions, as well as overall performance.
}
\label{train_freeze_comp}
\end{table*}


To assess the effect of three different training stages on model performance, we explore the following scenarios: 1) Direct Fine-tuning, where the model is exclusively trained on VQA-RAD or SLAKE 1.0 datasets without any prior training phases. 2) One-stage Pre-Training, which includes pre-training stage 1, followed by fine-tuning on downstream datasets. 3) Full Pre-Training, where the model undergoes all three training stages. This comparison offers valuable insights into the most effective training pathway for this model architecture in domain-intensive MedVQA tasks. 

Table \ref{train_stage_compare} shows the comparison results with BiomedCLIP-RadBloomz-7b. The findings reveal significant improvements in final accuracy, with an approximate 25\% increase in full pre-training (Scenario 3) compared to direct fine-tuning (Scenario 1). These results underscore the effectiveness of Pre-training stage 1, which greatly enhances the model's medical knowledge. Furthermore, full pre-training not only preserves the knowledge gained during stage 1 but also integrates medical concept alignment with specialized MedVQA training.


\begin{table}[t]
\centering
\resizebox{\linewidth}{!}{
\begin{tabular}{l|ccc}
\hline
Scenarios              & Overall  & Closed-ended & Open-ended \\ 
\hline
1      & 48.3                 & 59.9          & 30.7        \\
2          & 59.0                 & 70.6          & 41.3       \\
3 & 73.2 & 83.5 & 57.5 \\       
\hline
\end{tabular}
}
\caption{The table demonstrates the performance of our VLM (BiomedCLIP ViT+Radbloomz-7b) on VQA-RAD under different training scenarios: 1) direct fine-tuning on VQA-RAD; 2) stage 1 pretraining followed by fine-tuning on VQA-RAD; and 3) full pre-training and fine-tuning on VQA-RAD. The accuracy metric is used, and performance is reported for open-ended, closed-ended questions, along with overall accuracy.}
\label{train_stage_compare}
\end{table}

We examine the overall accuracy of VLMs using BiomedCLIP-ViT as the vision encoder across different question categories in both datasets. The results can be found in Tables \ref{category_compare_vqarad} and \ref{category_compare_slake}.
Our VLM with medical-tailored Radbloomz-7b shows better performance in most categories. 
RadBloomz-7b particularly excels in interpreting spatially-oriented queries, as evident from its leading performance in modality, abnormality, presence of objects/attributes, organ, and plane categories. This suggests a strong capability of RadBloomz-7b in analyzing the spatial arrangement in radiology images. However, the model can be further improved in shape, size, and position categories. Additionally, the distribution of categories in the training data has an impact on the model's performance.

\begin{table}[t]
\small \centering
\begin{tabular}{l|c|c|c}
\hline
Category & \#Q & Bloomz-7b1  &  RadBloomz-7b \\ 
\hline
Abnormality	& 56	& 64.3 &	\textbf{69.6} \\
Attribute	& 20	& 90.0	& 90.0 \\
Color	& 4	& 100.0 &	100.0 \\
Count	& 6	& 66.7	& \textbf{83.3} \\
Modality &	33	& 45.5	& \textbf{48.5} \\
Organ	& 10	& 20.0	& \textbf{40.0} \\
Plane	& 26	& 73.1	& \textbf{76.9} \\
Position	& 61	& 72.1	& 70.5 \\
Presence &	171	& 74.9	& \textbf{82.5} \\
Size	& 46	& 87.0	& 82.6 \\
Other	& 26	& 30.8	& 26.9 \\
\hline
\end{tabular}
\caption{Models’ overall accuracy (\%) across different question categories on VQA-RAD. Performance of two VLMs with Radbloomz-7b and Bloomz-7b1 as LLM component is reported separately. The vision encoder of VLMs is BiomedCLIP ViT. \#Q: number of questions in the given category.}
\label{category_compare_vqarad}
\end{table}

\begin{table}[t]
\small\centering
\begin{tabular}{l|c|c|c}
\hline
Category & \#Q & Bloomz-7b1  &  RadBloomz-7b \\ 
\hline
Organ	& 253 	& 88.9 &	\textbf{93.6} \\
Abnormality	& 	150 &  73.3	&  \textbf{84.6} \\
Size	&  65 & 86.1 &	\textbf{87.6} \\
Position	& 186 &	67.2 &  \textbf{87.6} \\
Plane  & 58 &	96.5	& \textbf{100.0} \\
Modality	& 108 &  100.0	& 100.0 \\
Knowledge Graph	& 148 &	68.9 & \textbf{75.0} \\
Color	& 34 &	88.2 & \textbf{91.1} \\
Quantity & 52 & 59.6 & 59.6 \\
Shape & 7 & 85.7 & 71.4 \\
\hline
\end{tabular}
\caption{Model’s overall accuracy (\%) across different question categories on SLAKE 1.0-English. Performance of two VLMs with Radbloomz-7b and Bloomz-7b1 as LLM component is reported separately. The vision encoder of VLMs is BiomedCLIP ViT. \#Q: the number of questions in the given category.}
\label{category_compare_slake}
\end{table}

\begin{table*}[t]
\centering
\begin{tabular}{c|l|l|l}
\hline
  & Question                  & Label         & Prediction \\ \hline
1  & What kind of image is this?        & x-ray                  & chest x-ray      \\
2  & What type of MRI sequence is displayed in this image?      & t2 weighted mri        & t2 weighted        \\
3  & What modality was used?            & plain film             & plain film xray  \\
4 & Are pleural opacities located on the left, right, or & & \\ 
& both sides of the lung? & both & both sides \\
5 &  Are there multiple or just 1 metastatic focus? & one & just one \\
6 & Which lung is clearer? & left & right \\ 
7 & Is the anatomy of the brain gyri affected? & no & yes \\ \hline
\end{tabular}
\caption{Examples of our model's generated answers (Prediction) on closed- and open-ended questions in VQA-RAD evaluated as incorrect answer.}
\label{qualitative}
\end{table*}

\begin{figure*}
\tiny \centering
\begin{subfigure}[b]{0.3\textwidth}
         \centering
         \includegraphics[width=2cm]{synpic31400.jpg}
         \caption{question 1}
     \end{subfigure}
     \begin{subfigure}[b]{0.3\textwidth}
         \centering
         \includegraphics[width=2cm]{synpic17738.jpg}
         \caption{question 4}
     \end{subfigure}
     \begin{subfigure}[b]{0.3\textwidth}
         \centering
         \includegraphics[width=2cm]{synpic100176.jpg}
         \caption{question 5}
     \end{subfigure}
\caption{Image examples from VQA-RAD corresponding to questions in Table \ref{qualitative}.}
\label{image_exmpl}
\end{figure*}

Finally, we conduct a qualitative analysis of the model's predictions to identify areas where improvements may be needed for both the model and evaluation measures. Table \ref{qualitative} shows examples of questions from the VQA-RAD test split where the model's predictions are evaluated as incorrect during the evaluation.
Notably, despite the model's responses being evaluated as incorrect according to our evaluation measure, a closer examination reveals a different perspective. The model provided responses that consist of terms that are either synonyms or contextually relevant to the given labels. For instance, in question 1, the model identifies the modality as `chest x-ray', which is essentially correct in the context of this question (See Figure \ref{image_exmpl}). Similarly, for question 2, the model's prediction `t2 weighted' captures the essence of the `t2 weighted mri' label or in question 4, `both sides' is predicted whereas the label is `both'. 

Given that traditional accuracy metrics may not fully capture the nuances and utilization of synonyms in the medical domain, conducting a manual evaluation of the predictions can be valuable in determining the actual performance of the model. However, it is worth noting that we have identified instances where the model generated incorrect answers, such as in questions 6 and 7. We asked a licensed medical expert to meticulously compare the model's predictions with the ground truth values and identify cases similar to those mentioned earlier. Following this rigorous human evaluation, we achieved an accuracy of $64.2\%$, surpassing the performance obtained using our automatic evaluation metric, which yielded an accuracy of $57.5\%$.



Although BiomedCLIP-RadBloomz-7b VLM demonstrates remarkable overall improvement in MedVQA, additional investigation of the model is necessary. Specifically, since the task is formulated as free-form generation, training a model to adhere to a restricted set of terminologies presents challenges and warrants further attention.

\section{Background and Related Work}

Language models (LMs) designed for general domains often face difficulties when applied to highly specialized fields. Additionally, data scarcity is a prevalent challenge in domain adaptation of LMs. Various methods have been developed to adapt pretrained LMs to specific domains. One method involves continuous pre-training of model parameters using data specific to the target domain \cite{radbloomz}. Alternatively, synthetic data can be effectively incorporated into the training process for fine-tuning models to better adapt to specific target domains \cite{karn-etal-2021-shot}. Another approach includes using parameter-efficient fine-tuning methods \cite{xu2023parameter} with task-specific training data. Our training schema amalgamates several of these methods like image-caption pretraining, synthetic biomedical MQA, and task-specific MedVQA datasets, all fine-tuned using a parameter-efficient technique.

Among parameter-efficient fine-tuning approaches, the Low-Rank Adaptation (LoRA) technique \cite{lora} has received considerable interest for adapting Large LMs (LLMs). In the biomedical domain, domain-specific LLMs have been proposed either by fine-tuning the model's parameters \citep{biogpt,pmcllama} or by utilizing LoRA techniques \citep{clinicalLlama}. However, it's important to note that biomedical domain-adapted LLMs might not perform as effectively in the radiology domain. This is due to the complexity of terminologies in clinical NLP \cite{karn-etal-2022-differentiable,ghosh-etal-2023-radling}. Thus, there have been recent proposals for radiology domain-adapted LLMs \citep{radbloomz}.

The application of domain adaptation is not limited to LLMs. It also finds utility in the adaptation of multi-modal models like vision-language models (VLMs). In line with this, there have been recent proposed biomedical VLMs such as \citep{biomedclip,pmcclip,medflamingo,chen2023medblip,llavamed}. These have been successful in achieving state-of-the-art performance in downstream biomedical tasks, such as medical question-answering. In this study, we concentrate on developing a more efficient domain adaptation technique for VLMs within the challenging domain of Radiology.

\section{Conclusion}

We introduce a new vision-language model for medical visual question-answering by integrating a radiology large language model, RadBloomz-7b \citep{radbloomz} and a biomedical vision encoder, BiomedCLIP-ViT \citep{biomedclip}, in to the VLM. 
Our main objective is to investigate the impact of integrating specialised LMs and vision encoders into VLMs for domain-specific tasks in the medical domain.

For this purpose, we propose a parameter-efficient training approach by deploying low-rank adaptation technique \citep{lora} to the decoder-only LLM component in the VLM, which significantly reduces the number of trainable parameters while maintaining the model performance. 
Moreover, the vision encoder is kept frozen in the training process. We then propose a two-stage pre-training approach aiming to align our VLM to medical concepts by pre-training the model on the image-captioning task and acquiring general knowledge for medical visual question answering by pre-training it on a general MedVQA dataset. We finally finetune the model on the downstream MedVQA tasks.

Our results demonstrate state-of-the-art performance on a MedVQA SLAKE 1.0 dataset and strong performance on the VQA-RAD dataset. 
Furthermore, compared to a VLM with a general-domain LLM, we show that our proposed VLM leads to a higher performance using parameter-efficient training, while a VLM with general-domain LM benefits slightly from training the vision encoder as well. Finally, our findings suggest that the proposed pre-training approach significantly improves model performance in downstream MedVQA tasks. 

\section{Limitations}

In this paper, we explored the generation ability of our adapted vision-language model on learning to generate free-form answers. While we observed impressive performance, we realized that in a few test cases, such as wh-questions, the model generates \textit{yes/no} answers. Therefore, more investigation on optimizing the training to capture the type of the question is required.

We proposed a multi-modal model tailored for radiology-domain visual-question answering tasks. Therefore, we are aware that our model is not easily generalizable to diverse medical domains and tasks, such as pathology image analysis. As a result, we didn't compare our model to SoTA generalized multi-modal models in other medical domains and tasks.
Furthermore, the LLM model architecture we studied is restricted to a decoder-only type, thus its performance may not be directly comparable to different model architectures.


\section{Ethics Statement}

All datasets in this paper are publicly available for clinical NLP research.
Trained models for MedVQA tasks in this paper must be assessed carefully before considering them for final applications.
\section{Disclaimer}
The concepts and information presented in this paper are based on research results that are not commercially available. Future commercial availability cannot be guaranteed.

\section{Acknowledgement}
We would like to sincerely thank the clinicians from the Siemens Healthineers Digital Technology and Innovation team for their support in the human evaluation of our results. Our appreciation also extends to the Siemens Healthineers supercomputing center for providing the necessary training infrastructure for our work. Furthermore, we acknowledge and appreciate the constructive feedback from anonymous reviewers, which significantly contributed to the refinement of our work.
\bibliography{acl_latex}

\appendix
\section{MedVQA datasets statistics}
\label{data_apn}

\begin{table*}[ht]
\centering
\caption{Downstream dataset statistics of VQA-RAD and SLAKE 1.0, includes number of images and question-answer pairs (QAs). Questions are categorized as close-ended and open-ended.}
\label{tab:datasets}
\begin{tabular}{l|rrr|rrrr}
\hline
\multirow{2}{*}{Dataset} & \multicolumn{3}{c|}{\textbf{VQA-RAD}} & \multicolumn{4}{c}{\textbf{SLAKE 1.0-English}} \\
                         & Total       & Train       & Test      & Total  & Train & Validation & Test \\ \hline
\#Images                 & 315         & 314         & 203       & 642    & 586   & 174        & 96   \\
\#QAs                    & 3515        & 3064        & 451       & 12995  & 9835  & 2099       & 1061 \\
\#Close-ended QAs               & 2093        & 1821        & 272       & 5141   & 3881  & 844        & 416  \\
\#Open-ended QAs                & 1420        & 1241        & 179       & 7754   & 5854  & 1255       & 645  \\ \hline
\end{tabular}
\end{table*}

\end{document}